## Fixed Suffix Dependency Ratio: Quantifying the Dual-Track Mechanism of Gender Assignment in Latvian Loanwords

Yelingyun Zhang[1,2]
[1] Faculty of Computer Science, Information Technology and Energy, Riga Technical University, Riga, Latvia
[2] Vidzeme University of Applied Sciences Scientific Institute, Valmiera, Latvia

Corresponding author: Yelingyun Zhang
Email: Yelingyun.Zhang@rtu.lv
ORCID: 0009-0002-7209-3587

### Abstract

Existing research has repeatedly observed the tendency for English loanwords to cluster in the masculine gender across different recipient languages, yet the origin of this pattern remains difficult to determine, as fixed morphological rules and default assignments are frequently analysed together. This study proposes the Fixed Suffix Dependency Ratio (FSDR) to quantify the degree of reliance on fixed derivational suffixes across different genders, and to distinguish between morphological anchoring and free-choice in distribution. By examining 1,832 Latvian noun lemma types, the results reveal a significant FSDR asymmetry within the loanword system: feminine loanwords rely significantly more on fixed derivational suffixes, while masculine loanwords are more concentrated in the free-choice zone. This pattern exhibits loanword specificity and has become more pronounced in contemporary usage. FSDR therefore provides a quantitative framework for testing default gender and shows how masculine default can be activated and reinforced under language contact.

### 1 Introduction

When loanwords enter a language with grammatical gender, lexical borrowing brings with it the problem of gender assignment. Cross-linguistic studies have repeatedly observed that nouns borrowed from English often tend to be assigned to the masculine gender in languages such as German, Polish and Spanish (Arndt, 1970; Poplack et al., 1982; Onysko, 2007; Fuchs, 2014; Balam, 2016). However, the source of this masculine bias is not always clear. It might originate from the automatic application of morphological rules already present in the target language, or from the default assignment path, or it might result from the combination of both within the same data. When all loanwords are pooled together, rule-driven assignment and default-driven assignment are mixed in the resulting proportions and cannot be separately identified.

This problem constitutes a core methodological gap in loanword gender research. Default or unmarked gender is often invoked to explain assignment tendencies when rule-based cues are insufficient, but without clear operational boundaries, it can also be read as a post-hoc label for residual phenomena. Depending solely on the masculine proportion among loanwords is insufficient to determine whether the masculine actually functions as the default assignment path. To verify this, it is necessary first to separate vocabulary constrained by fixed morphological rules from that which is not at the data level, then to examine whether the latter still exhibits systematic grammatical gender bias, and further to compare whether this bias is equally present in the native lexical system.

This study proposes the Fixed Suffix Dependency Ratio (FSDR) as a quantitative metric for enabling this separation. FSDR measures the degree to which a particular gender category relies on fixed derived suffixes. When one gender's loanwords rely more upon fixed-suffixes to maintain their assignment, whereas the other gender's loanwords are concentrated in areas not constrained by fixed-suffixes, this asymmetry reveals structural differences between morphological anchoring and default assignment. Different from the direct measurement of loanword masculine rates, FSDR reframes the question at the mechanistic level: how do different genders rely on fixed derivation rules; do loanwords and native words exhibit different distributions within the same morphological zone; and do these differences vary according to contact conditions?

Latvian language provides a highly suitable case study for this issue. Latvian features an active and clearly defined system of derivational suffixes, which makes it possible to draw a clear distinction between areas of fixed-suffixes and those of free choice. Meanwhile, our data reveal that the native lexical system does not exhibit a preference for masculine forms; the overall proportion of masculine forms is 44.6%, whilst in the free-choice zone that proportion stands at 42.4%. This suggests that if loanwords show a significant shift towards masculine forms within the same free-choice zone, this tendency cannot be simply explained as a reflection of the existing baseline in the native lexicon of the recipient language. Over the past century, Latvian has undergone a shift in its primary contact sources. Earlier gendered languages provided mappable gender cues for loanwords; the growing dominance of English has placed increasing numbers of new loanwords in an environment with no source-language gender information. This historical context makes Latvian a suitable language for examining changes in gender allocation mechanisms under conditions of gender vacuum.

Based on an analysis of 1,832 Latvian noun lemma types, this study finds a significant FSDR asymmetry in the loanword subsystem. The loanword FSDR Asymmetry is +0.336, showing that feminine loanwords depend on fixed derivational suffixes far more than masculine loanwords do. In the free-choice zone, loanword masculine proportion reaches 65.8%, compared to 42.4% for native words in the same zone, indicating that masculine default primarily manifests in the loanword subsystem. Diachronic comparison further shows that the loanword FSDR Asymmetry has increased from +0.215 in the Historical period to +0.395 in the Contemporary period, suggesting that this mechanism differentiation has intensified as contact conditions have changed. Suffix definition sensitivity tests and stem-final phonological control analyses support the robustness of this result.

The contributions of this study are primarily in two areas. Methodologically, FSDR provides a framework for separating fixed morphological rules from default assignment tendencies at the data level, enabling gender studies of loanwords to shift from comparisons of overall proportions to comparisons of mechanism-based dependencies. Theoretically, in a language whose native free-choice zone does not skew masculine, the fact that loan concepts systematically shift towards the masculine after entering the same morphological zone indicates that the masculine default is a structural phenomenon which is activated and reinforced under conditions of language contact.

The following sections first establish the theoretical basis for separating morphological rules from default mechanisms, then introduce data, suffix classification and the FSDR metric, followed by the presentation of empirical results, and finally the discussion on the implications of these findings for default gender, gender vacuum and cross-linguistic comparisons.

## 2 Theoretical Background

When a loanword enters the target language, the assignment of its grammatical gender involves two different mechanisms: morphological coercion and cognitive default. Current literature in this field often treats the two as a single entity in its analyses, which leads to the observed gender bias not being accurately attributed. This section aims to establish a theoretical basis for quantitatively distinguishing between these two mechanisms: Section 2.1 distinguishes the two mechanisms from a cross-linguistic perspective; Section 2.2 examines derivational suffixes as a diagnostic tool for isolating morphological coercion; Section 2.3 reviews cross-linguistic evidence from loanword research to clarify what observational conditions are needed; Finally, Sections 2.4 and 2.5 explain why Latvian constitutes an ideal subject for analysis in terms of both its morphological features and its diachronic development.

### *2.1 Gender assignment mechanisms: morphological coercion and cognitive defaults*

It is widely accepted within the field that the assignment of grammatical gender in a language is governed by multiple mechanisms (Corbett, 1991; Hellinger and Bußmann, 2003; Matras, 2020). Traditional frameworks view this as a hierarchical system, in which semantic criteria, such as biological sex, are applied first, followed by formal rules based on phonetic or morphological features, and if neither of these applies, a default gender is assigned (Corbett, 1991). Corbett (1991) hypothesises that when loanwords enter the target language, they are assigned gender in accordance with the normal rules of assignment, just like other nouns. This means that the assignment of gender to nouns is a layered process in which different types of information are given different levels of priority.

When considered from the perspective of language contact, this competitive mechanism becomes even more pronounced, with the gender assignment of loanwords competing against phonological analogy, semantic association and unmarked default options in relation to native vocabulary (Poplack et al., 1982). Furthermore, when phonological and semantic cues conflict or are absent, the default gender is more readily activated. Several cross-linguistic studies have found that masculine serves as the default gender, and that the gender of loanwords tends to align with it (Arndt, 1970; Poplack et al., 1982; Balam, 2016; Fuchs, 2014; Onysko, 2007). Poplack et al. (1982) also found that nouns borrowed from English remain subject to the inflectional and gender systems of the recipient language, though in empirical statistical analysis, the precise observable boundary between this rule-based coercive assignment and default assignment remains unclear.

The explanatory power of the concept of default gender itself has long been questioned within the field, Kilarski (1997) referred to it as a ‘dustbin category’ lacking substantive meaning, arguing that any observed gender bias in loanwords can in fact be explained by the normal gender assignment rules of the recipient language. The critique exposes a methodological dilemma: unless words whose gender is determined by morphological rules are examined separately from those that reflect the cognitive choices of native speakers, any observed bias cannot be accurately attributed.

To overcome this dilemma, it is essential to clarify the boundaries of morphological rules. Harris (1991) points out that suffixation and gender are two independent systems that merely happen to be correlated. This distinction opens up the possibility that, if certain derivational suffixes fundamentally determine gender rather than just being associated with it, they constitute an independent, high-priority assignment mechanism that takes precedence over the default rule (Fraser and Corbett, 1995; Onysko, 2007; Thornton, 2009).

Yet in empirical studies of loanword gender assignment, morphological coercion and cognitive default have not been separated at the data level. Existing work reports overall gender proportions for loanwords as a whole, treating suffix-determined cases and freely assigned cases as a single pool. This statistical confound might mask the true extent of cognitive default, thus establishing a quantitative metric capable of measuring reliance on morphological rules could prove key to resolving this entanglement.

### *2.2 The role of derivational suffixes in the gender system*

When discussing the distinction between morphological rules and free allocation, it is important to consider the properties of derivational suffixes within the lexicon. Cross-linguistic research has shown that nominalising morphology imposes distributional requirements on gender (Kramer, 2020), and that gender-marking suffixes are pre-assigned specific gender features in the lexicon (Doleschal, 2015). For example, in German, the suffix *-in* consistently derives feminine nouns, whereas the agentive suffix *-er* derives masculine nouns, this correspondence between morphology and gender constitutes an assignment rule. When a noun carries these suffixes, its gender is predetermined by the suffix and is no longer subject to open assignment. The gender of such derived words is a direct outcome of derivation itself. This deterministic role creates a binary partition in the lexicon: words whose gender features are fully specified by a derivational affix, and those in which the gender is unspecified, words whose gender must be resolved through syntactic agreement or default rules (Ralli, 2002). This partition provides the theoretical basis for the present study's distinction between fixed-suffix nouns and free-choice nouns.

The influence of morphological rules is also reflected in cognitive processing. Whether morphological cues are transparent or reliable also affects the pathways through which gender information is extracted. Research by Bates et al. (1995) shows that transparent gender markers can speed up gender extraction in Italian. Taraban and Roark (1996), on the other hand, demonstrated through a competitive model that cue reliability is a key predictor of gender learning. When morphological cues reliably indicate a particular gender, competition is low and allocation is efficient. Conversely, competition intensifies, and the system tends to rely on statistical patterns to carry out allocation.

In terms of language evolution from a macro-level perspective, the long-term stability of gender systems depends on a rich set of morphological consistency markers (Audring, 2014), whilst the transition from transparent to non-transparent morphology is a common diachronic trajectory (Wälchli and Di Garbo, 2019). English, for example, underwent precisely this erosion, reducing its gender system to pronoun-level natural-sex distinctions (Curzan, 2003, as cited in Audring, 2014). In systems where gender classification has not yet been completely lost, morphological erosion can also shift the weights of the original rules, as seen in northern Scandinavian languages, weakening of word-final morphology has been accompanied by gender shifts in specific word classes (Van Epps et al., 2021). Under conditions of incomplete acquisition, morphologically opaque nouns are the first to lose their original gender marking, with speakers defaulting to masculine (Polinsky, 2008).

This morphological dependency also extends to the process of integrating new words, such as in the Baltic language family, where morphological adaptation is the primary strategy for integrating loanwords. In Lithuanian, derivational suffix replacement serves as a core mechanism for creating native equivalents, and the gender carried by the suffix overrides other assignment cues (Pakerys, 2016). Derivational suffixes, then, are not passive maintainers of the gender system—they also function as active assigners during loanword nativization.

These properties of derivational suffixes offer a new angle on Kilarski's (1997) critique of default gender. Kilarski argued that default gender lacks explanatory power because all assignment can be traced back to normal rules. But if derivational suffixes operate coercively within the system, their assignment outcomes will be folded into overall statistics, obscuring whatever other mechanisms may be at work. This also means that derivational suffixes can serve as a diagnostic tool. By separating out the words whose gender is determined by coercive derivational morphology, the remaining words are freed from that morphological constraint. If these words continue to exhibit a systematic gender bias, and if this bias is absent in native terms but appears only in loanwords, it is difficult to attribute this to internal syntactic constraints of the target language; rather, it is more likely to reflect an independent cognitive default tendency on the part of speakers when morphological cues are lacking.

### *2.3 Loanword gender in contact linguistics*

At the lexical level, nouns are the most frequently borrowed word class, making up the majority of foreign vocabulary in bilingual corpora (Pakerys, 2016; Balam, 2016; Matras, 2020). Considering that nouns are also the core word class that carries grammatical gender, the assignment of gender in loanwords is an essential morphological issue in the study of language contact. Previous research has suggested that the grammatical gender of loanwords may be influenced by a variety of factors, including the natural gender, the grammatical gender in the source language, the grammatical gender of the equivalent term in the target language, and the morpho-phonological characteristics of the loanword itself (Matras, 2020). In practice, these factors interact, and the gender of any single loanword is rarely predictable from one variable alone.

When the source language and the target language have similar gender assignment systems, loanwords can directly adopt the gender of the source language, but when gender-neutral English enters a language with grammatical gender, the absence of morphological cues from the source language forces the target language to rely on its own assignment mechanisms (Levkovych, 2024). Cross-linguistic studies have repeatedly observed this tendency toward a masculine default in such contexts: after English loanwords entered German, the early practice of word-for-word analogy gradually shifted toward a mechanism in which masculine serves as the default gender (Onysko, 2007); the masculine default strategy for English loanwords in Polish also shows a trend toward expansion (Fuchs, 2014); in mixed phrases in Northern Belizean Spanish, the rate of masculine default usage even reaches 87% (Balam, 2016). This evidence from different language families together demonstrates that, when the source language lacks gender cues, the masculine default is a systematic cross-linguistic tendency rather than a coincidental phenomenon specific to individual languages.

Suffix-based analogy runs parallel to this default tendency. Arndt (1970), studying German bilinguals, found that when suffix analogy cues are present, they dominate assignment and produce highly consistent results. Boers et al. (2020) observed in Dutch-Spanish bilinguals that default strategies and translation-equivalent analogy strategies operate simultaneously. These studies all indicate that the default strategy is not the only allocation pathway, with morphological analogy and cognitive default serving as two parallel strategies that correspond to the fixed-suffix zone and the free-choice zone examined in this study.

One methodological difficulty in existing research, however, is that in most languages studied so far, native vocabulary itself skews masculine. In Russian, masculine nouns make up 46% of the noun lexicon (Corbett, 1991). In German, high-frequency nouns split roughly 40% masculine, 30% feminine, and 30% neuter (Arndt, 1970). Similar patterns hold in Spanish and Polish (Harris, 1991; Stefańczyk, 2007), where masculine dominates in both loanwords and native words. This consistency across languages obscures whether the masculine bias is caused by contact or by the distributional characteristics of the language itself. Therefore, a language that does not exhibit a masculine bias in its native system would provide clearer observational conditions for distinguishing between these two sources.

### *2.4 The Latvian morphological and gender system*

Latvian is an inflected language with two grammatical genders—masculine and feminine—and no neuter form. Its noun system has a symmetrical structure: six declension classes, three masculine (first through third) and three feminine (fourth through sixth) (Kalnača and Lokmane, 2021). Inflectional endings largely signal gender. Consonant-final stems typically enter masculine declension classes, stems ending in *-a* or *-e* are mostly feminine. But the mapping is not absolute, nouns from different declension classes can share the same surface ending in particular case forms while belonging to different genders, for example, the genitive ending *-a* of a first-declension masculine noun is identical in form to the nominative ending *-a* of a fourth-declension feminine noun. In terms of gender distribution, among the 1,302 local lexical items collected, the overall gender distribution of Latvian local nouns differs from that of most previously studied languages: feminine accounts for 55.4% and masculine for 44.6% (see Section 4.4 for details), showing no significant tendency toward a masculine default.

For the purposes of this study, another key feature of the Latvian morphological system lies in the relationship and rules governing derivational suffixes and gender. The official reference grammar confirms that derivational rules remain highly productive in modern Latvian (Kalme and Smiltniece, 2001). Following the morphological classification in Kalnača and Lokmane (2021), suffix–gender relationships fall into two types (for the full suffix inventory, see Section 3.2):

- Single-gender suffixes, where gender is fully determined by the suffix itself. For example, *-šan-* (action nouns), *-īb-* (abstract concepts), and *-tav-* invariably produce feminine nouns, while *-um-* (result nouns) invariably produces masculine nouns.
- Dual-gender suffixes, where gender is determined not by the suffix alone but by the natural sex of the referent. Agentive *-tāj-*, characterizing *-ul-*, and professional *-niek-*, among others, have paired masculine and feminine forms.

In addition to the two categories mentioned above, there are also nouns in Latvian that do not carry derivational suffixes. Based on this morphological feature, this study divides nouns into two categories when examining the mechanisms of gender assignment, thereby providing a classificatory basis for subsequently isolating default tendencies from the data. Nouns with derivational suffixes are classified as fixed-suffix, since both single-gender and dual-gender suffixes assign gender through explicit rules rather than open choice; nouns without such suffixes are classified as free-choice, their gender is not constrained by derivational morphology and constitutes the primary zone for observing default mechanisms.

This distinction is not limited to Latvian. Across languages with grammatical gender, this logic of separation is transferable, as derivational suffixes denoting gender are widespread in such languages (Corbett, 1991; Fraser and Corbett, 1995; Ralli, 2002; Doleschal, 2015; Pakerys, 2016; Kramer, 2020; Van Epps et al., 2021). What Latvian adds is an observational advantage: its native system does not skew masculine. If a masculine bias appears in the loanword subsystem, it can be more cleanly attributed to contact-induced cognitive default rather than to an inherent property of the receiving language.

***2.5 The diachronic context: from isomorphic mapping to gender vacuum***

The sources of loanwords in the Latvian language have experienced a structural shift over the past century. In the first half of the 20th century, German and Russian were the main contact languages, both of which feature a grammatical gender system (Rūķe-Draviņa, 1977). Given this context, the grammatical gender of the source language provides clues for assignment, and loanwords can be transferred to the target language via an isomorphic mapping of the source language's gender. For example, German *Karte* 'card' (F) and *Fakt* 'fact' (M) were borrowed into Latvian as *karte* 'card' (F) and *fakts* 'fact' (M) respectively. By the end of the 20th century, following Latvia's restoration of independence, English had replaced Russian as the primary language of contact and mediation for Latvian, bringing not only a wide range of direct loanwords but also influencing Latvian word-formation patterns and expression habits (Veisbergs, 2018). At present, approximately 70 per cent of the Latvian language corpus consists of translated texts, and translation is becoming a major driving force behind the evolution of the language (Veisbergs, 2009).

Unlike in earlier source languages, English vocabulary lacking grammatical gender provides no clear morphological clues when borrowed into Latvian, creating a 'gender vacuum'. As the influence of English vocabulary on European languages has grown, this shift in characteristics has transformed the path of integration for loanwords (Görlach, 2002). When the conditions for isomorphic mapping no longer hold, the disappearance of morphological cues is accompanied by the reorganisation of the assignment mechanism. As discussed in Section 2.2 regarding the morphological weakening effect, in the North Scandinavian languages, feminine words exhibit marked instability following the weakening of phonological cues, with loanword status being a significant predictor of gender instability (Van Epps et al., 2021). From a macro perspective, gender systems also generally tend to become less transparent over time (Wälchli and Di Garbo, 2019). The default mechanism's role is amplified in interactions where gender cues are absent—in such contexts, receiving language relies more upon its own allocation mechanism to assign gender to these words.

In the context of linguistic theory, this shift towards reliance on default mechanisms is often regarded as a sign of system simplification. Intensive language contact typically leads to a reduction in morphological complexity (Trudgill, 2011), whilst the blurring of morphological boundaries is even regarded as an early sign of linguistic structure decline (Dorian, 1981). However, testing this hypothesis of decline requires the ability to distinguish in empirical analysis between overall trends and localised responses under specific conditions of contact. Given this shift in Latvian loanwords, if it becomes possible to distinguish empirically between words governed by morphological rules and those relying on default assignment, and to compare the diachronic differences exhibited by native and loanwords in these two zones, the actual impact of the gender vacuum on the language will be quantified more clearly. From a structural perspective, this provides a new angle for analysing the mechanisms of language evolution under contact pressure.

## 3. Data and methods

To operationalize the separation of morphological coercion from cognitive default proposed in Section 2, this section translates the theoretical framework into a concrete quantitative design. It describes the dataset used (Section 3.1), the suffix classification criteria and the basis for partitioning nouns into fixed-suffix and free-choice zones (Section 3.2), and the definition of the Fixed Suffix Dependency Ratio (FSDR) along with the statistical framework (Section 3.3).

### *3.1 Dataset and annotation*

This study has constructed a dataset comprising 1,832 Latvian noun lemma types, including 530 loanwords and 1,302 native words. As this study focuses on integration mechanisms at the lexical type level rather than frequency of use within texts, all statistical analyses are based on unique lemma types rather than token frequency. The corpus data were drawn from three sources:

- A publicly available, manually annotated dataset extracted from Latvian Wikipedia (Zhang et al., 2025), with high inter-annotator agreement (Cohen's $\kappa > 0.83$).
- Open-access educational materials from the Skolo.lv platform, operated under the Skola2030 initiative of Latvia's National Centre for Education (VISC). Three subject areas at the secondary level were selected: *Sociālās zinātnes* ‘social sciences’, *Dizains un tehnoloģijas* ‘design and technology’, and *Dabaszinības* ‘natural sciences’.
- A historical expansion consisting of three Latvian-language texts published before 1985 and corresponding to the same subject areas: *Sociālā psiholoģija un vēsture* ‘Social Psychology and History’ (Poršņevs, 1982), *Latvijas dizains* ‘Latvian Design’ (Ancītis et al., 1984), and *Latvijas zeme, daba un tauta* ‘Latvia’s Land, Nature and People’ (Malta and Galenieks, 1936).

Taking into account the shift in the sources of Latvian loanwords described above, the dataset was divided into two groups based on time period: the Historical group (up to and including 1985) and the Contemporary group (from 1985 onwards). The Historical group comprises 654 lexical items (171 loanwords and 483 native words), and the Contemporary group (which includes data from Wikipedia and textbooks) comprises 1,178 lexical items (359 loanwords and 819 native words).

In the data pre-processing and annotation workflow, this study employed deep learning classifiers and natural language processing methods. To avoid distortion from global deduplication across time periods, deduplication was performed within each period separately: if the same lemma appeared in both subsets, it was retained once in each. For loanword identification, the study used an mBERT binary classifier fine-tuned on Latvian-language data from a bert-base-multilingual-cased checkpoint. The model achieved a loanword classification F1 of 0.899 on a held-out test set (Zhang et al., 2026). For part-of-speech filtering and gender extraction, the Stanza Latvian pipeline was used to tokenize, POS-tag, and lemmatize the data. Only items tagged as NOUN (UPOS) were retained, and their grammatical gender labels (Gender=Masc or Gender=Fem) were extracted following Universal Dependencies conventions. At the same time, in order to verify the reliability of the automatic annotation model, this study conducted stratified random sampling from the data annotated by the model (n=100, with 25 samples per period and per label), and Latvian language researchers carried out a blind manual review to verify the two dimensions of loanword/native word labelling and gender annotation. The loanword classification achieved a Macro-F1 of 0.93 in this validation sample (loanword Precision = 0.960, Recall = 0.906; native-word Precision = 0.900, Recall = 0.957). In this validation dataset, no errors were found in the noun gender features extracted by Stanza, with an accuracy rate of 100%. This indicates that the NLP pipeline has achieved the accuracy required for empirical research regarding gender determination in Latvian.

### *3.2 Suffix classification and matching procedure*

The classification criteria for suffixes in this study are primarily based on the list of Latvian derivational systems provided by Kalnača and Lokmane (2021), the validity of which has also been confirmed in the official reference grammar of Latvian (Kalme and Smiltniece, 2001). As outlined in the theoretical framework of Section 2.4, nouns are partitioned into two operational categories: fixed-suffix and free choice.

The fixed-suffix category covers all nouns whose gender assignment is constrained by an explicit morphological or semantic rule. The full inventory used in this study contains 51 derivational suffixes, organized into three types:

- 15 single-gender suffixes (feminine): *-šan-, -īb-, -tav-, -tuv-, -oņ-, -ain-, -otn-, -atn-, -av-, -m-, -sm-, -nīc-, -ij-, -itāt-, -ācij-*
- 10 single-gender suffixes (masculine): *-um-, -āj-, -ēn-, -tēn-, -lēn-, -uk-, -ism-, -īv-, -ekl-, -ukl-*
- 26 dual-gender suffixes (gender determined by the natural sex of the referent): *-tāj-, -ul-, -el-, -tel-, -niek-, -iniek-, -eniek-, -on-, -iņ-, -tiņ-, -utiņ-, -sniņ-, -īt-, -iet-, -n-, -tn-, -sn-, -ēj-, -kl-, -okl-, -l-, -sl-, -ien-, -en-, -ist-, -isk-*

As discussed in Section 2.4, the common feature of these three types of suffixes is that gender assignment is not influenced by the speaker's free choice, but is determined by morphological or semantic rules. In contrast, free-choice category covers all nominal morphological contexts that do not carry the aforementioned derivational suffixes. In such contexts, the gender of a noun is not determined by derivational forms or natural gender rules, but is influenced by factors such as phonological default, semantic analogy or cognitive default.

For the actual data matching, the study uses an automated identification procedure. Inflectional endings are first stripped from each lemma to obtain the derivational stem. The stem is then matched against the known suffix list using a longest-match-first strategy to avoid false matches on shorter suffixes. For single-character suffixes, a minimum root length of four characters is required to reduce false positives. In this study, the results of the automatic recognition were manually reviewed by conducting a stratified random sample from the annotated data (n = 200, comprising 50 fixed zone and 50 free zone loanwords, and 50 fixed zone and 50 zone free native words), with Latvian language researchers verifying the accuracy of suffix recognition and classification on a case-by-case basis. The review results showed an overall accuracy rate of 97.5 per cent (195/200), comprising three false positives and two false negatives. These errors do not affect the direction or significance of the results reported in this study. To assess whether results depend on how suffix boundaries are drawn, three definition methods were also tested as a sensitivity check. The first uses only the baseline list from Kalnača and Lokmane (2021). The second adds data-driven candidates: suffixes appearing at least 10 times with a purity of 95% or higher, where purity is the proportion of cases in which the suffix co-occurs with the same gender. The third extends the frequency threshold to 5 or more. Results under all three definitions are reported in Section 4.3.

### *3.3 The Fixed Suffix Dependency Ratio (FSDR) and statistical framework*

To quantitatively separate morphological coercion from cognitive default in language contact, this study proposes the Fixed Suffix Dependency Ratio (FSDR). For a given gender G, FSDR is defined as the conditional probability of carrying a fixed derivational suffix:

$$FSDR_G = P(Fixed \mid G)$$

where Fixed denotes the event that a noun falls into the fixed-suffix category. The metric measures how much a given gender depends on derivational morphology for its assignment. The definition generalizes to any number of grammatical genders. To capture the structural difference between genders, the study further defines an Asymmetry index. For Latvian's two-gender system:

$$Asymmetry = FSDR_{Fem} - FSDR_{Masc}$$

When Asymmetry is positive, this indicates that, within the linguistic system, the feminine form relies more on the support of fixed-suffixes than the masculine form, whereas the masculine form is assigned more through cognitive default mechanisms that are not constrained by morphology. For languages with three or more grammatical genders, the gender that is least dependent on suffixes can be identified as the default candidate by comparing FSDR values in pairs.

The theoretical basis of the mathematical definition of FSDR lies in the fact that it operationalises the two allocation mechanisms discussed earlier into a measurable quantity: the difference between morphological compulsion, corresponding to the fixed-suffix zone, and cognitive default, corresponding to the free-choice zone. Different from traditional methods, which report gender preferences as an overall frequency, FSDR reveals the respective contributions of morphological rules and default tendencies within the gender system by comparing differences in the degree of reliance on fixed-suffixes across genders.

The empirical analysis proceeds in four steps: first, testing for masculine default in the free-choice zone; second, calculating the FSDR asymmetry; third, comparing asymmetry between loanwords and native words; and fourth, examining whether this difference changes across time periods. The stability of the Asymmetry estimate is assessed via bootstrap resampling (10,000 iterations, 95% confidence interval), if the interval excludes zero, the asymmetry is considered significant. Whether the two assignment zones produce different gender patterns is tested with a chi-square independence test. A binomial test assesses whether the masculine proportion among free-choice loanwords deviates significantly from the 50% baseline.

To test whether FSDR asymmetry is jointly modulated by gender, borrowing status, and time period, a logistic regression model is fitted with fixed-suffix status as the dependent variable (1 = fixed-suffix domain, 0 = free-choice domain). The model was specified as:

$$\text{is_fixed} \sim \text{gender} * \text{borrowing_status} * \text{period}$$

Where the two-way interaction gender $*$ borrowing_status tests whether the FSDR asymmetry is specific to loanwords; the three-way interaction tests whether this asymmetry has increased in the contemporary period. Finally, to rule out stem-final phonology as an alternative explanation for the masculine bias in the free-choice zone, a supplementary phonological control analysis is conducted. This analysis compares the entropy of stem-final consonant distributions across masculine and feminine nouns, calculates the correlation of per-consonant masculine rates between native words and loanwords, and fits a logistic regression controlling for stem-final consonant to test whether borrowing status independently predicts masculine assignment. All data processing and statistical analysis in this study were carried out using Python, primarily utilising the pandas, numpy and scipy libraries.

## 4. Results

In this section, the FSDR framework proposed earlier is applied to Latvian noun data to examine the differing roles of fixed derivation rules and free selection zones in the gender assignment of loanwords. Section 4.1 compares the gender distributions of loanwords and native words across both zones and calculates the FSDR asymmetry, asking whether masculine default appears only among loanwords in the absence of fixed derivational constraints. Section 4.2 uses a diachronic comparison to test whether this mechanism dependence has intensified as contact conditions have changed. Section 4.3 then assesses the robustness of the results through suffix-definition sensitivity tests and a stem-final phonological control analysis.

### *4.1 Gender distribution and FSDR asymmetry*

Figure 1 presents the gender distribution of loanwords and native words in the fixed-suffix zone and the free-choice zone (exact counts in Table 1). If the fixed derivation rules and the free-choice zone do in fact correspond to two different mechanisms of gender assignment, then the two should exhibit different patterns in terms of gender proportions. And if masculine default is a contact-specific phenomenon, the difference should appear primarily in the loanword subsystem.

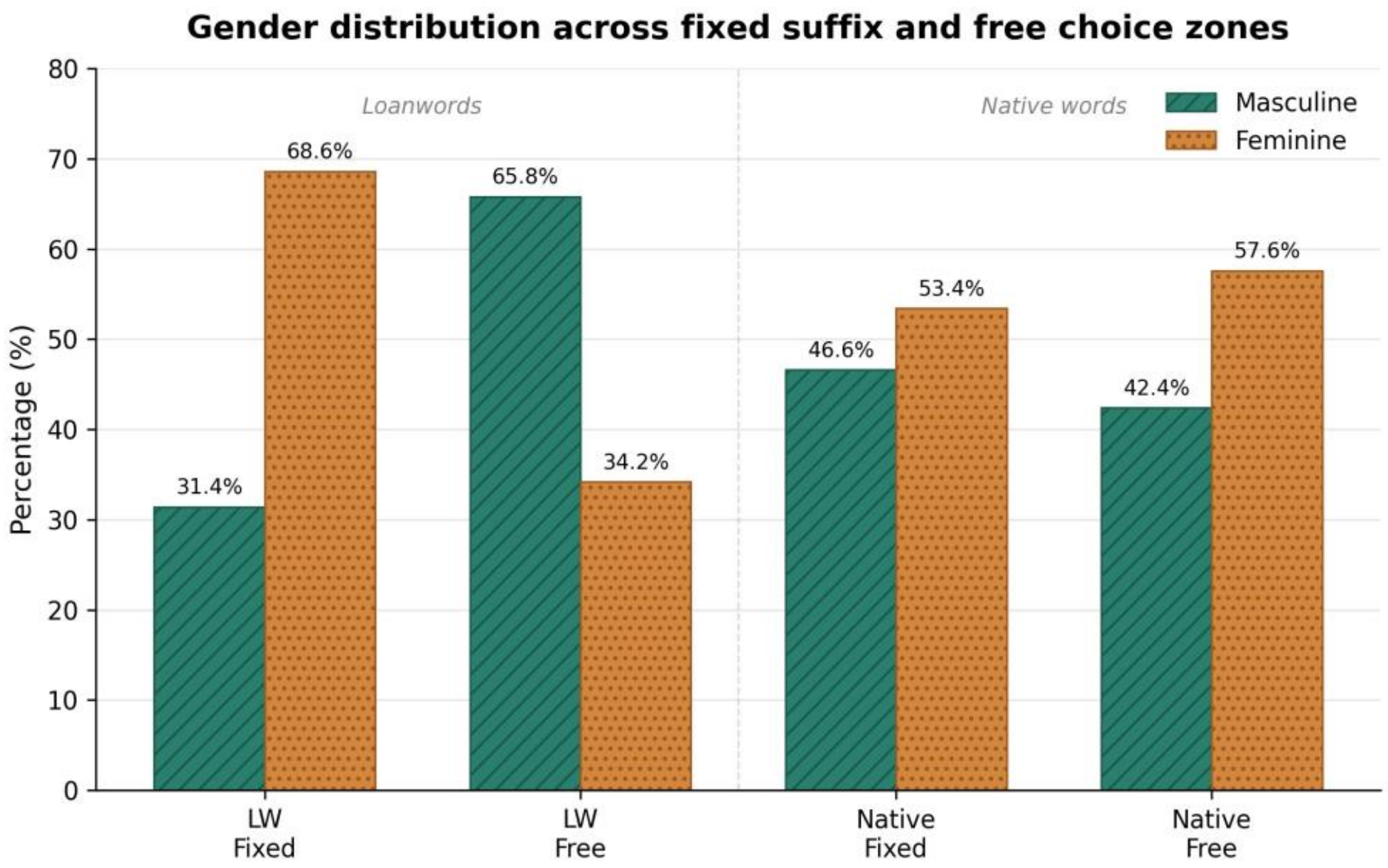

*Fig. 1. Gender distribution of loanwords and native words across fixed-suffix and free-choice zones.*

Table 1. Gender distribution in fixed-suffix and free-choice zones.

| Group | Total N | Fixed: Masc | Fixed: Fem | Free: Masc | Free: Fem |
|---|---|---|---|---|---|
| Loanword | 530 | 31.4% (71) | 68.6% (155) | 65.8% (200) | 34.2% (104) |
| Local word | 1302 | 46.6% (322) | 53.4% (369) | 42.4% (259) | 57.6% (352) |

Between the two allocation zones, loanwords exhibit a marked reversal in gender proportions. Within the fixed-suffix zone, only 71 of the 226 loanwords are masculine, representing 31.4%; in contrast, there are 155 feminine loanwords, accounting for 68.6%. This distribution indicates that fixed derivational suffixes primarily support the retention and generation of feminine gender in loanwords. For example, loanwords such as *konsultācija* 'consultation' (*-ācij-*) and *aktivitāte* 'activity' (*-itāt-*) contain fixed feminine suffixes and are therefore assigned to the fixed-suffix zone; likewise, *dekorējums* 'decoration' (*-um-*) contains a fixed masculine suffix and is classified on the basis of derivational morphology. The gender assignment of these words is governed by identifiable derivation rules, and thus stands in contrast to loanwords, which are not subject to fixed-suffix constraints.

In the free-choice zone, the gender distribution of loanwords shifts towards the masculine. Of the 304 loanwords without a fixed-suffix, 200 were masculine and 104 were feminine, representing a proportion of 65.8% masculine. Compared with the group of words with fixed-suffixes, the proportion of masculine words increased by 34.4%. Examples of freely chosen positive loanwords include *projekts* 'project', *transports* 'transport' and *produkts* 'product'; by contrast, free-choice feminine loanwords include *kultūra* 'culture', *mašīna* 'machine' and *taktika* 'tactics' and so on. These internal variations within the group indicate that the surface gender ratio of loanwords does not directly reflect a single mechanism. Where derivational constraints are present, feminine forms predominate; when such constraints disappear, the proportion of masculine forms rises rapidly.

This free-choice masculine bias does not appear in native words. The proportion of masculine forms among local words was 46.6% (322/691) in the fixed-suffix zone and 42.4% (259/611) in the free-choice zone, with neither zone showing a predominance of masculine forms. The difference between loanwords and native words in the free-choice zone is highly significant ($\chi^2 = 43.53$, $p = 4.17 \times 10^{-11}$). This comparison serves as evidence confirming that this positive default is not a general feature of the native Latvian nominal system, but rather becomes evident only after loanwords enter this area, where there are no fixed derivational constraints. As a supplementary reference, the masculine proportion among free-choice loanwords also significantly exceeds the 50% baseline (binomial test, $p = 3.95 \times 10^{-8}$).

Through FSDR, the aforementioned differences in distribution are further converted into gender-specific fixed-suffix dependency. Among loanwords, feminine shows a markedly higher dependence on fixed-suffixes: FSDRFem = 0.598 (155/259), while FSDRMasc = 0.262 (71/271). Nearly 60% of feminine loanwords depend on fixed-suffixes for their gender assignment; for masculine loanwords, the figure is roughly 26%. The resulting FSDR Asymmetry is +0.336 (bootstrap 95% CI [0.254, 0.414]). The confidence interval does not include zero.

Local terms, on the other hand, exhibit a completely different pattern. Among these, the dependence on fixed-suffixes for masculine and feminine forms was almost equal, with FSDR_Masc at 0.554 and FSDR_Fem at 0.512, corresponding to an asymmetry value of just –0.042. This suggests that, within the native lexical system, the two genders exhibit a broadly symmetrical degree of reliance on fixed-suffixes, and there is no structure characterised by a high degree of feminine dependency, as is found in the loanword subsystem. A bootstrap difference test confirms that the gap between loanword and native-word asymmetry is significant (95% CI [0.281, 0.476]).

Logistic regression supports the same conclusion. With fixed-suffix status as the dependent variable, the gender * borrowing status interaction is significant (coef = 0.951, p = 0.010), indicating that the gender gap in suffix dependence is significantly larger among loanwords than among native words. Borrowing status amplifies the structural difference between genders. It can thus be seen that the FSDR captures a structural asymmetry arising during the integration of loanwords, which differs from the general gender distribution in the Latvian noun system, where feminine loanwords rely more on fixed derivational forms to maintain their proportion in the distribution, whilst masculine loanwords are more concentrated in areas without fixed derivational constraints.

### *4.2 Diachronic change in FSDR asymmetry*

Between the two periods examined in this study, the FSDR asymmetry of loanwords showed a clear increase. The asymmetry value for the historical period (up to and including 1985) was +0.215. By the contemporary period (from 1985 onwards), the asymmetry value had increased to +0.395, representing a rise of approximately 84% (Figure 2; see Table 2 for period-level values). This change results from the combined effect of two factors: the FSDR for masculine loanwords fell from 0.302 to 0.243, indicating that masculine loanwords are less concentrated in the fixed-suffix zone; the FSDR for feminine loanwords, on the other hand, rose from 0.518 to 0.638, indicating that feminine loanwords are more concentrated in the fixed-suffix zone.At the same time, the proportion of masculine loanwords in the free-choice zone also increased from 59.4% (60/101) to 69.0% (140/203), indicating a higher degree of concentration in this zone. Together, these two factors contributed to an increase in the FSDR asymmetry value.

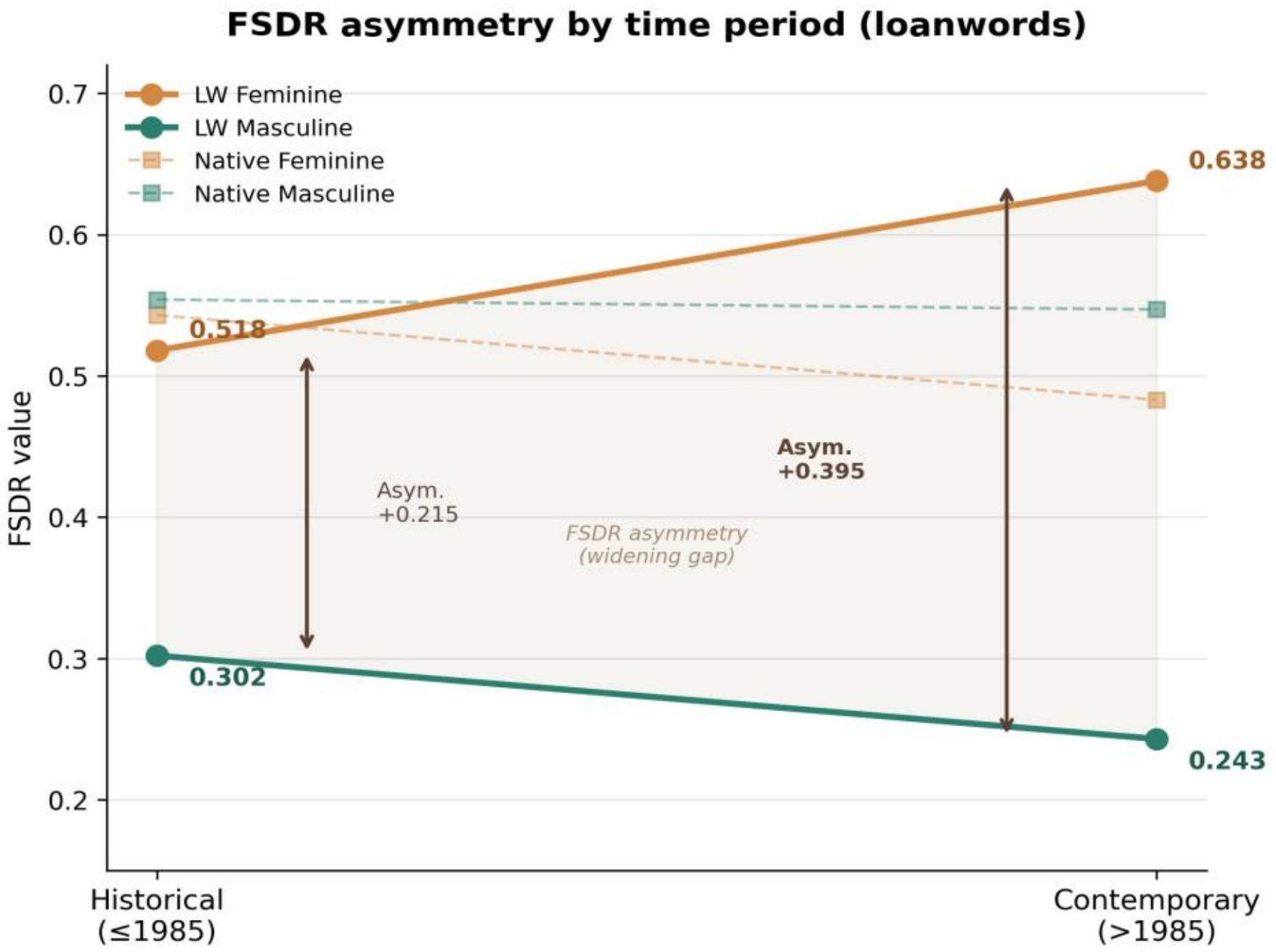

*Fig. 2. Diachronic change in loanword FSDR by gender, with native word baselines.*

Table 2. FSDR asymmetry by time period.

| Period | N (LW) | Masc FSDR | Fem FSDR | Asymmetry | Free Masc% (n) |
|---|---|---|---|---|---|
| Historical (≤1985) | 171 | 0.302 | 0.518 | +0.215 | 59.4% (60/101) |
| Contemporary (>1985) | 359 | 0.243 | 0.638 | +0.395 | 69.0% (140/203) |
| All periods | 530 | 0.262 | 0.598 | +0.336 | 65.8% (200/304) |

This approach is consistent with the shift in contact conditions discussed earlier. Among the sources of loanwords throughout history, languages with grammatical gender such as German and Russian have played a significant role; the gender of the source language can provide clues for mapping onto the target language. By contrast, contemporary loanwords are more influenced by English. The absence of grammatical gender in English renders gender cues at the source language level unavailable. Under these conditions, the target language’s internal derivation rules and default mechanisms jointly play a more significant role in gender assignment. The diachronic changes in FSDR indicate that this shift has widened the structural differences between feminine and masculine in terms of their reliance on fixed-suffixes.

Unlike loanwords, native words do not present the same diachronic pattern. Among the 483 local words from the historical period, the FSDR asymmetry value was –0.011; among the 819 local words from the contemporary period, this value was –0.064. The asymmetry values for local words in both periods are close to zero, indicating that the degree to which masculine and feminine forms rely on fixed-suffixes remains broadly symmetrical within the local word system. It is evident that, in the type-level data analysed in this study, the diachronic increase in FSDR asymmetry occurs primarily within the loanword subsystem, and no corresponding systematic changes are observed in native words.

Statistical tests support this diachronic trend. A bootstrap difference test on the Asymmetry values yields a 95% CI of [0.008, 0.357], which does not include zero. A logistic regression with fixed-suffix status as the dependent variable shows a significant three-way interaction of gender * borrowing status * period (coef = 1.007, p = 0.028), confirming that the gender gap in suffix dependence among loanwords has widened in the Contemporary period. A supplementary proportions z-test on the free-choice masculine rate shows a directionally consistent but weaker trend (59.4% vs. 69.0%, z = 1.655, p = 0.098). Together, these results support the conclusion that the FSDR structural asymmetry has intensified over time, consistent with the strengthening of masculine default under modern contact conditions.

### *4.3 Robustness and control analyses*

Having confirmed the asymmetry of FSDR in loanwords and its diachronic enhancement, this study conducted two supplementary analyses to examine whether the main findings were dependent on specific methodological boundaries or latent formal factors.

Because FSDR depends on the partition between fixed-suffix and free-choice zones, the study tested three suffix definition methods to check whether the core results are sensitive to where the boundary is drawn (Table 3). Although the scale of loanwords in the suffix-fixed zone has expanded, the FSDR asymmetry values remain positive under all three definitions. The masculine proportion in the free-choice zone fluctuates slightly but stays consistently above 60%. The dual-track structure captured by FSDR is not an artifact of a particular suffix boundary. Furthermore, after removing duplicate lexemes across time periods (n=1540), the asymmetry value of loanword FSDR remained unchanged (+0.336), indicating that the results are also insensitive to the deduplication strategy.

Table 3. FSDR stability across suffix definitions.

| Definition | Fixed Masc | Fixed Fem | Free Masc | Free Fem | Asymmetry | Free Masc% |
|---|---|---|---|---|---|---|
| Strict baseline | 71 | 155 | 200 | 104 | +0.336 | 65.8% |
| Threshold ≥10, purity ≥95% | 136 | 173 | 135 | 86 | +0.166 | 61.1% |
| Threshold ≥5 | 154 | 199 | 117 | 60 | +0.200 | 66.1% |

Another aspect to consider is that the masculinebias in the free-choice region may also be influenced by the phonological structure at the end of the word stem. In Latvian, stem-final consonants are associated with gender to some degree, so the analysis must rule out the possibility that loanwords simply follow the same phonological gender habits as native words. Table 4 summarizes three complementary phonological control results.

Table 4. Phonological control analysis in the free-choice zone.

A. Entropy of stem-final consonant distributions.

| Sample | Masc entropy | Fem entropy | Fem − Masc |
|---|---|---|---|
| Local words, all | 2.313 | 2.638 | +0.325 |
| Local words, free choice | 2.507 | 2.942 | +0.435 |
| Loanwords, free choice | 2.048 | 2.416 | +0.368 |

B. Correlation of per-consonant masculine rates.

| Comparison | Shared consonants | r | p |
|---|---|---|---|
| Local all vs. local free choice | 21 | 0.585 | 0.005 |
| Local free choice vs. loanword free choice | 12 | 0.270 | 0.395 |
| Local all vs. loanword free choice | 12 | −0.120 | 0.710 |

C. Phonological predictive power (logistic regression $R^2$).

| Model sample | $R^2$ | p |
|---|---|---|
| Local words, all | 0.232 | $1.01 \times 10^{-74}$ |
| Local words, free choice | 0.142 | $2.29 \times 10^{-15}$ |
| Loanwords, free choice | 0.368 | $4.02 \times 10^{-22}$ |

As can be seen from Table 4-A, the entropy values for masculine nouns are lower than those for feminine nouns across the entire sample of native words, the free-choice zone for native words, and the free-choice zone for loanwords. This means that in Latvian, masculine is concentrated in fewer stem-final consonant types, while feminine is spread more broadly. Therefore, at the phonological level, Latvian does not exhibit a default masculine gender. If measured in terms of the scope of consonant application, the feminine gender actually has a wider distribution. This proves false the explanation that attributes the masculine bias in loanwords to the native phonological system's default.

Table 4-B shows that loanwords do not follow native phonological patterns. There was a moderate correlation between the entire sample of native words and the free-choice zone of native words ($r = 0.585$, $p = 0.005$), indicating that even after removing fixed suffixes, a certain phonological–gender mapping structure remains within the native words. However, the correlation between the free-choice zone for native words and that for loanwords was not significant ($r = 0.270$, $p = 0.395$) and even showed a weak negative correlation between the entire sample of native words and the free-choice zone for loanwords ($r = -0.120$, $p = 0.710$). This indicates that the relationship between consonants and gender in loanwords is not a continuation of the phonological conventions of the native language.

This misalignment is also evident in specific consonant clusters. Stems ending in *-t-* constitute the largest group of consonants in the data. Among native free-choice words, *-t-* stems show 58.7% masculine (61/104); both masculine nouns like *pamats* 'foundation' and feminine nouns like *grāmata* 'book' are common. Among loanword free-choice words, the same *-t-* stems show 93.5% masculine (87/93). A similar reversal appears for *-k-*: 72.7% masculine in native words, 7.7% in loanwords. These results indicate that the consonant–gender mapping in loanwords follows its own pattern and cannot be traced back to the phonological distribution within the native lexicon.

Overall, the stem-final consonants do indeed play a role in gender assignment (see Table 4-C): the phonological model explains 23.2% of gender variance in native words overall, 14.2% in native free-choice words, and 36.8% in loanword free-choice words. This shows that the assignment of gender to loanwords is not entirely independent of the phonological structure; however, as can be seen from Table 4-B, this structure does not replicate the patterns found in native words, but rather forms a stronger and more extreme consonant–gender correspondence within the loanword subsystem.

In further regression models, where the final consonant of the word stem was included as a control variable, loanword status remained a significant predictor of masculine assignment. The controlled model showed that the effect of loanword status was independent (OR = 2.18, $p = 4.92 \times 10^{-6}$). Even when the influence of the final consonant of the stem is taken into account, the loanword status itself still significantly increases the probability of a masculine assignment. Such results further support the central argument of this study: that the masculine bias in the free-choice zone is an independent allocation tendency within the loanword subsystem, rather than a by-product of fixed derivation rules or local phonological structures.

## 5. Discussion

As can be seen from the results presented above, the overall tendency towards masculine gender assignment in Latvian loanwords reflects a dual-track structure comprising both fixed derivational rules and a free-choice zone. This section further explores the theoretical implications of this finding. By situating FSDR within the methodological debate surrounding default gender, it also examines the morphological dependence and diachronic reinforcement of feminine loanwords under conditions of a gender vacuum. Subsequently, it illustrates from a cross-linguistic perspective why FSDR provides a more robust comparative framework than absolute masculinity rates. Finally, it outlines the limitations of this study and suggests avenues for future research.

### *5.1 Quantifying the dual-track mechanism and empirically testing default gender*

By measuring FSDR, this study transforms the debate on default gender from a conceptual issue into a testable empirical question. In research on the gender assignment of loanwords, default or unmarked gender has long occupied an ambiguous position. On the one hand, it can explain allocation tendencies that cannot be directly determined by semantic, phonological or morphological rules; on the other hand, in the absence of clear operational boundaries, it can easily become a residual label used to explain remaining examples. It is in this sense that Kilarski (1997) criticises unmarked gender, arguing that it risks becoming a dustbin category lacking in descriptive and explanatory power. Enger (2009) further points out that, whilst this criticism may be somewhat strong, the default gender is indeed easily taken overly for granted, and therefore needs to be tested on the basis of evidence from within the language itself.

This debate points to a more fundamental question: how should default gender be empirically tested? Reporting only the overall proportion of positive forms among loanwords does not allow determining whether this bias stems from fixed morphological rules, phonological analogy, or a default tendency arising from the absence of rules. FSDR was developed precisely to address this observational dilemma. By separating nouns governed by fixed derivation rules from those without such suffixes, ensuring that the allocation results directly governed by morphological rules are no longer statistically conflated with those in open allocation zone, and by comparing the degree to which different genders rely on fixed suffixes, FSDR transforms the potential manifestation of the default mechanism into measurable structural differences. Building on this, the study further employs native words as an internal baseline, ensuring that the evaluation of the default mechanism does not rely on external presets, but is instead based on comparable references within the same linguistic system. If positive bias is merely the result of general gender assignment rules in the language, then, once fixed derivational suffixes are excluded, loanwords and native words should exhibit similar distributions. The results of this study show that, in Latvian, the two categories exhibit a significant divergence in the free-choice zone: the proportion of masculine loanwords reaches 65.8%, whilst that of native words in the same zone stands at 42.4%. This discrepancy suggests that the masculine bias cannot be directly attributed to the standard distribution pattern of the native Latvian noun system.

Therefore, as a new quantitative tool, FSDR examines the default phenomenon in the gender distribution of loanwords for the first time using mechanistic segmentation and internal controls. By separating the fixed-suffix zone from the free-choice zone, default gender is transformed from an abstract theoretical construct into a localisable phenomenon. The results address the methodological concerns raised by Kilarski and Enger: even after explicitly partitioning default gender and controlling for internal baselines and alternative explanations, systematic effects remain significant. More importantly, this effect does not follow the existing gender distribution patterns found in native words, but manifests as an independent distribution tendency within the free-choice zone of loanwords. When loanwords and native words both fall within a range not constrained by fixed derivational suffixes, they exhibit opposite gender distributions. Native words do not show a predominance of masculine forms, whereas loanwords show a significant concentration of masculine forms. This suggests that the default gender in Latvian is not a leftover category retrospectively labelled by researchers, but rather a contact-induced effect that persists even after known morphological rules and local phonological conventions have been ruled out. From this perspective, the concern with the default mechanism is not simply whether it is set too high, but rather the opposite, that its impact may be underestimated in standard aggregate statistics.

The fixation of derived suffixes provides morphological support for a large number of feminine loanwords, making the overall gender ratio of loanwords appear relatively balanced, however once these rule-bound words are set aside, the masculine default within the area of free choice becomes clearly apparent. The gender vacuum further amplifies this structure. When the primary source of contact shifts to English, which lacks grammatical gender, the pressure to assign gender is transferred more to the structure of the target language itself, thereby further magnifying the contrast between fixed derivational rules and areas of free choice. The increased FSDR asymmetry observed in the time-series results demonstrates that the masculine default gains a clearer and stronger sphere of influence under conditions of a gender vacuum.

This framework has also shifted the understanding of gender bias in loanwords. What FSDR reveals is a proportional imbalance, as well as the issue of the default categorisation of new words by gender when they enter the target language. The data from Latvian makes this particularly clear. These differences in grammatical gender by default reflect which category is regarded as the more natural, more fundamental and less in need of additional marking when rules are absent, explanations are insufficient or choices are open. FSDR asymmetry value of +0.336 quantifies precisely this inequality, which has long been obscured by the overall proportions: feminine loanwords must rely more upon morphological rules to maintain their position, yet masculine loanwords can gain a default advantage when no fixed rules apply. Thus, FSDR not only measures morphologically dependent variations in the gender assignment of loanwords, but also reveals how default gender is activated, amplified and reorganised within the loanword subsystem under conditions of language contact. When a language in a locally free-choice zone that initially exhibits non-masculine tendencies shifts towards masculine defaults upon the introduction of foreign concepts into that same open zone, this phenomenon may no longer be purely a technical aspect of gender assignment, but rather a structural phenomenon worthy of further investigation at the intersection of syntax, cognition and culture.

### *5.2 Cross-linguistic perspective and the methodological contribution of FSDR*

From a cross-linguistic perspective, one of the insights offered by this study is that the rate of masculine default in loanwords cannot be directly compared without taking into account the gender baseline of the target language itself. Previous research has observed a tendency for English loanwords to cluster towards the masculine gender in languages such as German, Polish and Spanish, but the gender distributions within the native lexical systems of these languages do not follow the same pattern. If only the proportion of absolute masculine forms among loanwords is compared, it is difficult for researchers to determine whether a high masculine rate in a given language reflects a stronger contact-induced default, or simply perpetuates the masculine bias already present in the recipient language.

FSDR provides a within-language comparison logic. It requires first distinguishing fixed derivational rules from the free-choice zone within a single language, then comparing loanwords and native words in the same zone, and then comparing how much each gender depends on fixed suffixes. Default effects are identified through structural differences between loanwords and native words, allowing contact-induced masculine default to be separated from the receiving language's own baseline. The analytical logic underpinning this measurement framework is also transferable across languages. For languages with different gender systems, different lists of derivational suffixes, different native lexical bases and different histories of language contact, provided that the language in question has fixed rules governing gender assignment, it is possible to calculate the degree to which each gender relies on these rules, and to further compare the asymmetrical differences between loanwords and native words. Therefore, the methodological contribution of FSDR also lies in its ability to be used to compare and measure how the gender assignment of loanwords in different languages deviates from the baseline of the native system.

### *5.3 Limitations and future research*

This study has the following limitations. Firstly, the study relies on written language materials, comprising mainly textbooks, reference texts and encyclopaedic corpora; consequently, the results reflect a relatively formal and standardised form of written Latvian. The gender assignment of loanwords may follow different patterns in spontaneous speech, social media texts and contexts involving code-switching; particularly in environments where normative pressure is weaker and real-time processing is more pronounced, the role of default mechanisms may become more directly apparent. Secondly, the analysis in this paper is conducted at the lemma type level. This approach is suitable for examining the gender distribution patterns of loanwords as they enter the lexicon, but it does not allow us to determine whether certain high-frequency loanwords exert a stronger systemic influence in actual usage. Future research could further examine the impact of frequency, discourse and context on the stability of gender distribution at the token level. Furthermore, although this study has ruled out several key alternative explanations through suffix sensitivity tests and phonological control analysis, there remain potential competing factors within the free choice domain, such as lexical domain, syllabic structure, the influence of equivalent terms in translation, and the degree of terminological standardisation. While these factors do not undermine the core findings of FSDR, the free-choice zone itself remains an area worthy of further investigation. Future research could analyse in greater detail how the masculine default is activated, amplified or suppressed under different exposure conditions.

## 6. Conclusion

This study proposes the Fixed Suffix Dependency Ratio (FSDR) as a means of distinguishing, at the data level, between fixed morphological rules and default assignment tendencies in the gender assignment of loanwords. Based on an analysis of 1,832 Latvian noun lemma types, this paper demonstrates that the gender assignment of loanwords cannot be fully explained by overall gender proportions; only by distinguishing between regions governed by fixed derivational suffixes and those governed by free choice can a structural division of labour between morphological anchoring and gender default be observed.

This study reveals that loanwords in Latvian exhibit a clear dual-track mechanism. The asymmetry value of FSDR for loanwords is +0.336, and the confidence interval does not include zero, indicating a significant difference between negative and positive words in their reliance on fixed suffixes. Feminine loanwords rely more strongly on derived forms to secure a stable position, whilst masculine loanwords are more likely to be assigned to areas without fixed derivational constraints. This structure is not a general feature of the native Latvian noun system. The fact that loanwords and native words exhibit significant differences within the same free-choice domain suggests that the masculine default is a contact-related phenomenon activated within the loanword subsystem. Suffix sensitivity tests and phonological control analyses further confirm that this result cannot be attributed to suffix boundary choices or native stem-final consonant habits.

The study also finds that FSDR asymmetry has intensified in contemporary loanwords. As gender cues in the source language have diminished, diachronic findings indicate that mechanistic differentiation within the loanword subsystem is increasing, with no corresponding change observed in the native lexicon. This suggests that contact pressure primarily affects the gender assignment patterns of loanwords as they enter the language, rather than the Latvian noun gender system as a whole.

The contribution of this study is to turn default gender from a concept debated in theoretical terms into an empirical object that can be quantitatively tested. FSDR reveals structural differences that overall gender proportions may obscure. Future research can apply FSDR to Lithuanian, Polish, and other gendered contact languages, comparing the degree to which loanwords deviate from native baselines across different systems. Further work can also incorporate additional factors to examine more precisely how default gender is activated, amplified, or suppressed under different contact conditions.

## Acknowledgements

This research was supported by the Institute of Digital Humanities, Faculty of Computer Science, Information Technology and Energy at Riga Technical University. The authors also acknowledge Marina Platonova and Atis Kapenieks for their supervision and guidance, and Chengming Ke for insightful discussions and feedback on earlier drafts of this paper.

## List of Abbreviations

FSDR — Fixed Suffix Dependency Ratio
LW — Loanword(s)
NLP — Natural Language Processing
mBERT — multilingual BERT (bert-base-multilingual-cased)
UPOS — Universal Part-of-Speech (tag)
UD — Universal Dependencies
OR — Odds Ratio
CI — Confidence Interval
VISC — *Valsts izglītības satura centrs* 'National Centre for Education Content'
F1 — F1 score (harmonic mean of precision and recall)